\def\year{2022}\relax

\documentclass[letterpaper]{article} 
\usepackage{aaai22}  
\usepackage{times}  
\usepackage{helvet}  
\usepackage{courier}  
\usepackage[hyphens]{url}  
\usepackage{graphicx} 
\usepackage{natbib}  
\usepackage{caption} 
\DeclareCaptionStyle{ruled}{labelfont=normalfont,labelsep=colon,strut=off} 
\usepackage{algorithm}
\usepackage{algorithmic}
\usepackage{subcaption}
\usepackage{multirow}
\usepackage{adjustbox}
\usepackage{caption} 
\usepackage{multirow}
\usepackage{scrextend}
\usepackage{multicol}
\usepackage{xcolor}

\usepackage{comment}
\usepackage{newfloat}
\usepackage{listings}
\floatstyle{ruled}
\newfloat{listing}{tb}{lst}{}
\floatname{listing}{Listing}

\title{Electricity Consumption Forecasting for Out-of-distribution Time-of-Use Tariffs}
\author{Jyoti Narwariya\equalcontrib, Chetan Verma\equalcontrib, Pankaj Malhotra, Lovekesh Vig,\\Easwara Subramanian, Sanjay Bhat}
\affiliations{
(jyoti.narwariya, verma.chetan, malhotra.pankaj, lovekesh.vig, easwar.subramanian, sanjay.bhat)@tcs.com\\
TCS Research, New Delhi, India
}

\usepackage{bibentry}
\usepackage{amsmath}
\usepackage{amsfonts}

\begin{document}

\maketitle
\begin{abstract}
In electricity markets, electricity retailers or brokers want to maximize profits by allocating tariff profiles to end-consumers. One of the objectives of such demand response management is to incentivize the consumers to adjust their consumption so that the overall electricity procurement in the wholesale markets is minimized, e.g. it is desirable that consumers consume less during peak hours when cost of procurement for brokers from wholesale markets are high.
We consider a greedy solution to maximize the overall profit for brokers by optimal tariff profile allocation, i.e. allocate that tariff profile to a consumer that maximizes the profit w.r.t. that consumer. This in-turn requires forecasting electricity consumption for each user for all tariff profiles. This forecasting problem is challenging compared to standard forecasting problems due to following reasons: i. the number of possible combinations of hourly tariffs is high and retailers may not have considered all combinations in the past resulting in a biased set of tariff profiles tried in the past, i.e. retailer may want to consider new tariff profiles that may achieve better profits, ii. the profiles allocated in the past to each user is typically based on certain policy, i.e. tariff profile allocation for historical electricity consumption data is biased. 
These reasons violate the standard i.i.d. assumptions as there is a need to evaluate new tariff profiles on existing customers and historical data is biased by the policies used in the past for tariff allocation.
In this work, we consider several scenarios for forecasting and optimization under these conditions. We leverage the underlying structure of how consumers respond to variable tariff rates by comparing tariffs across hours and shifting loads, and propose suitable inductive biases in the design of deep neural network based architectures for forecasting under such scenarios. More specifically, we leverage attention mechanisms and permutation equivariant networks that allow desirable processing of tariff profiles to learn tariff representations that are insensitive to the biases in the data and still representative of the task. Through extensive empirical evaluation using the PowerTAC simulator, we show that the proposed approach significantly improves upon standard baselines that tend to overfit to the historical tariff profiles. 
\end{abstract}

\begin{figure}[H]
   \captionsetup[subfigure]{font=small,labelfont=small, justification=centering}
    \centering
        \includegraphics[scale=0.5,trim={0.15cm 8.5cm 2.8cm 0.8cm},clip]{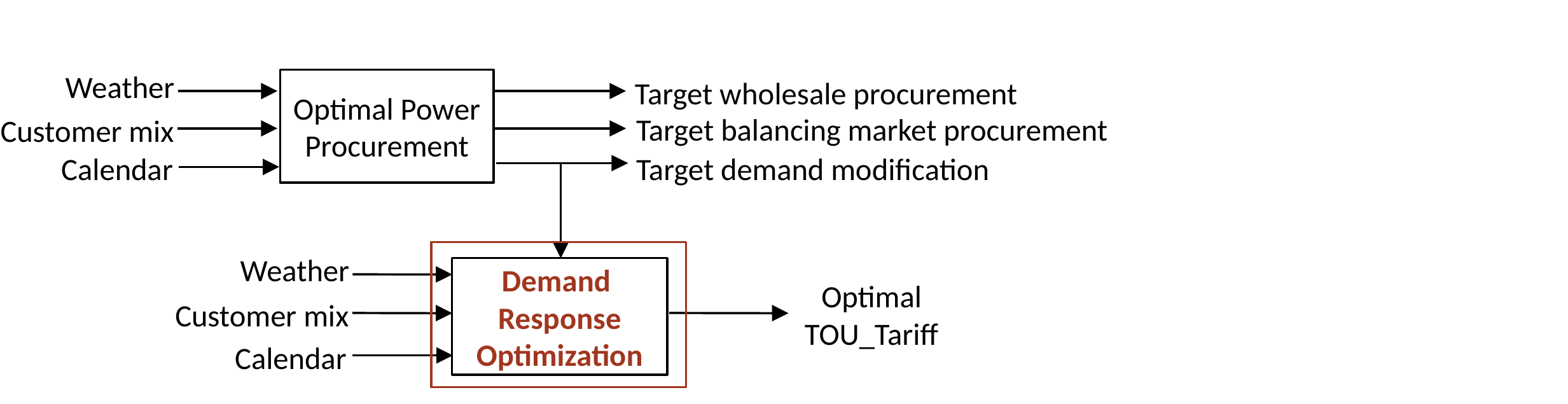}
        \caption{Various aspects and objectives in an electricity markets. In this work, we focus on a sub-problem related to allocation of optimal time-of-use tariff (TOU Tariff) to each customer\label{fig:elecmarket}}
\end{figure}

\section{Introduction}
A smart grid consists of multiple types of entities such as those involved in generation, distribution, and consumption (smart appliances and buildings).
One of the aims of a smart grid is to manage electricity demand in an economical manner via integration and exchange of information about all entities involved.
For the customers or the end-consumers as well as the electricity distributing agencies or the electricity brokers, it offers the flexibility to choose/allocate among dynamically changing tariffs to meet certain objectives, e.g. minimize electricity bill for customers, maximize profit for retailers, etc. However, meeting such objectives is challenging due to dynamics of the market, e.g. changing wholesale electricity prices, supply-demand fluctuations, etc.

As depicted in Figure \ref{fig:elecmarket}, a broker typically performs three functions: i. purchase or sell power to its subscribers or customers in the retail market, ii. purchase or sell power in the wholesale market, iii. rectify any supply-demand imbalance within its portfolio through the balancing market.
In this work, we consider a simplified setting where the broker performs following two functions: i. sell power to those customers in the retail market who are electricity consumers, and ii. purchase power in the wholesale market. 
Typical examples of consumers include offices, housing complexes, hospitals and villages. 
Furthermore, we focus on only those subset of consumers who have a \textit{shiftable} load component in their total or aggregate consumption in addition to the traditional fixed or non-shiftable load, i.e. the consumption (e.g. appliance usage) at an hour that cannot be moved to another hour. This shiftable load can be shifted from the originally preferred hour to another hour in the day if the tariff for the latter is lower. The broker may want to encourage such a behavior, known as demand response management \cite{siano2014demand}, to maximize profit or balance demand-supply.

In this work, we consider the following out-of-distribution generalization problem: given historical aggregated consumption of consumers to tariff profiles allocated to them, forecast the aggregated consumption for new tariff profiles. These new tariff profiles are part of the electricity broker or retailer's plan to explore new profiles to further improve the profits. This is different from standard forecasting problems as the exogenous variables (tariff profiles) at test time are different from the exogenous variables at train time. Furthermore, the allocation of tariff profiles in the past is not random, so the data is biased in the sense that for different consumer personas not all historical tariff profiles would have been tried.
We note that the logic based on which the consumers respond to tariff profiles is consistent irrespective of the tariff profile. We propose to capture that logic in the neural network by using permutation equivariant networks and attention mechanisms. 

Key contributions of this work can be summarized as follows:
\begin{itemize}
    \item We consider the problem of electricity consumption forecasting under new tariff profiles not encountered previously. This is then used for tariff profile allocation to optimize electricity broker's profits.
    \item We note that the forecasting problem can be seen as an out-of-distribution (OOD) generalization problem with bias in the training data consisting of temporal and confounding bias.
    \item To achieve OOD generalization, we leverage the logic behind how consumers respond to tariff profiles in order to shift load, and propose a novel neural network architecture to achieve better OOD generalization.
\end{itemize}
Through empirical evaluation, we show that the proposed approach is able to improve upon vanilla methods that do not take into account suitable inductive biases guided by the knowledge of how consumers respond to tariff profiles.

\section{Problem Formulation\label{sec:probdef}}
\begin{figure*}[h!]
   \captionsetup[subfigure]{font=small,labelfont=small, justification=centering}
    \centering
    \begin{subfigure}{0.42\textwidth}
    \centering
    \captionsetup{justification=centering}
    \includegraphics[width=10.6cm, trim={0.8cm, 5.0cm, 1.0cm, 0.2cm}]{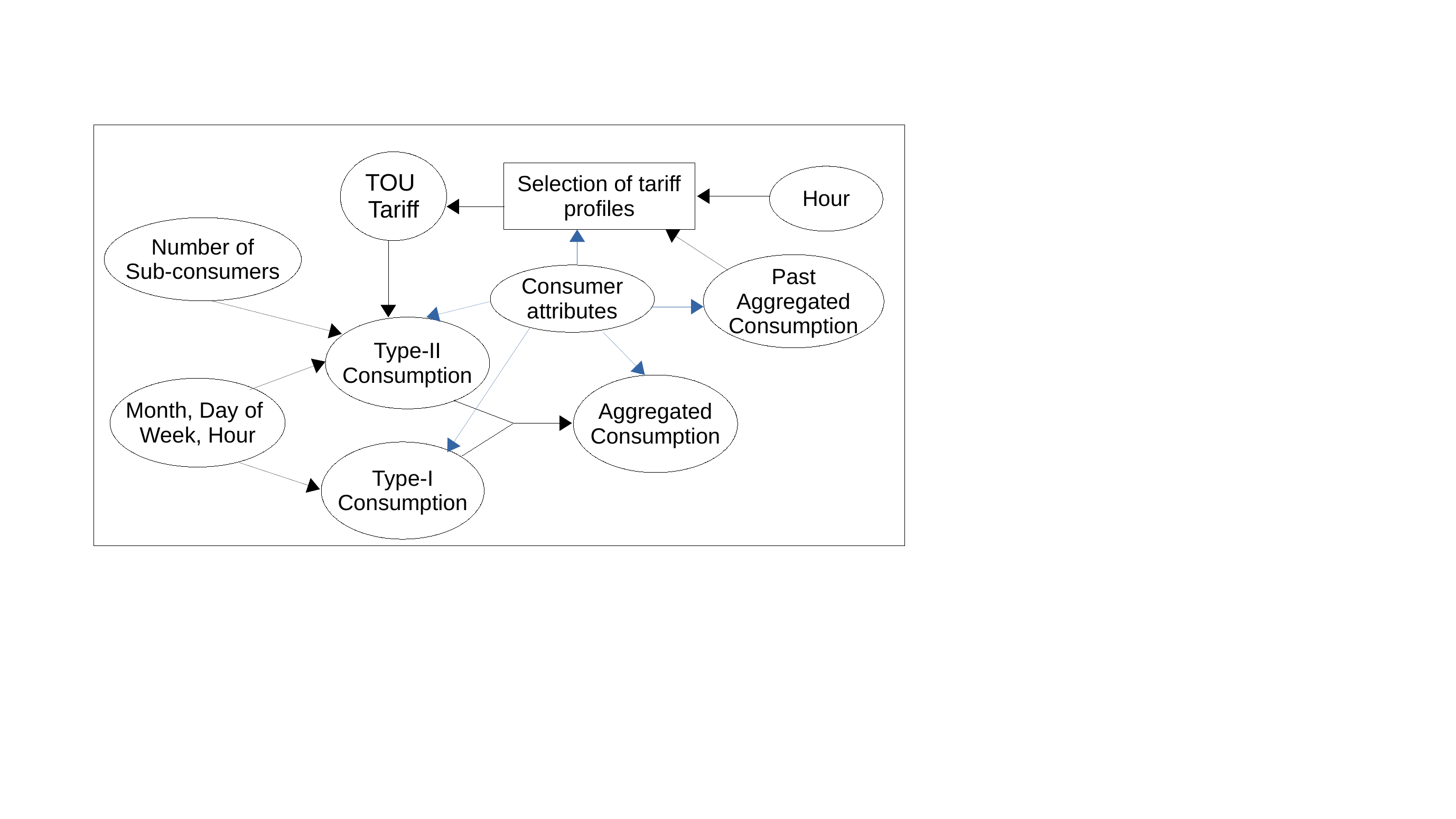}
    \caption{\label{fig:causaldiag}Causal Diagram}
    \end{subfigure}
    \begin{subfigure}{0.28\textwidth}
        \centering
        \includegraphics[scale=0.29,trim={0.6cm 0.5cm 0.2cm 0.2cm},clip]{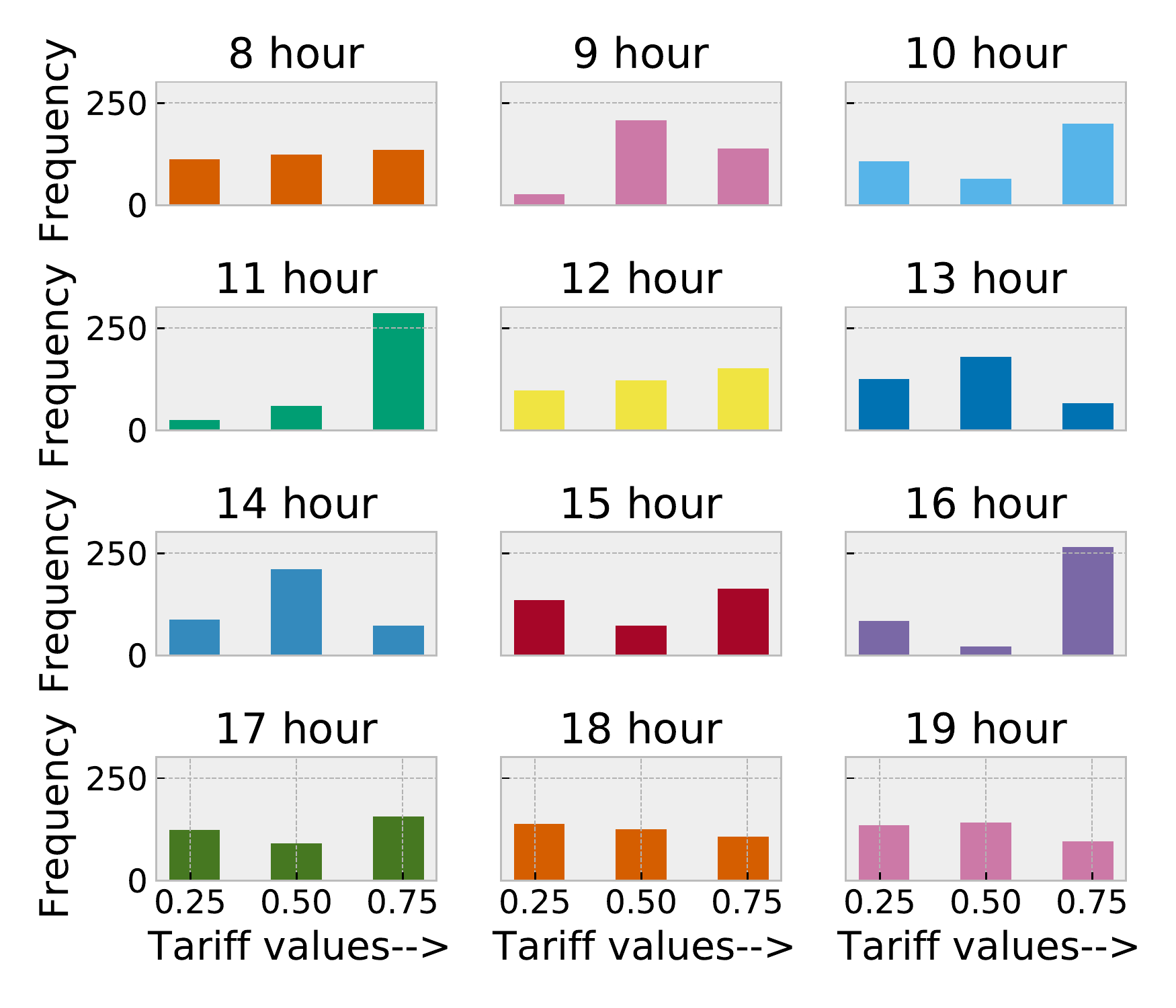}
        \caption{Hourly Tariff Distributions in IID Profiles depicting temporal bias ($\mathcal{T}_{in}$).}
    \end{subfigure}
    \begin{subfigure}{0.28\textwidth}
        \centering
        \includegraphics[scale=0.29,trim={1.15cm 0.5cm 0.2cm 0.2cm},clip]{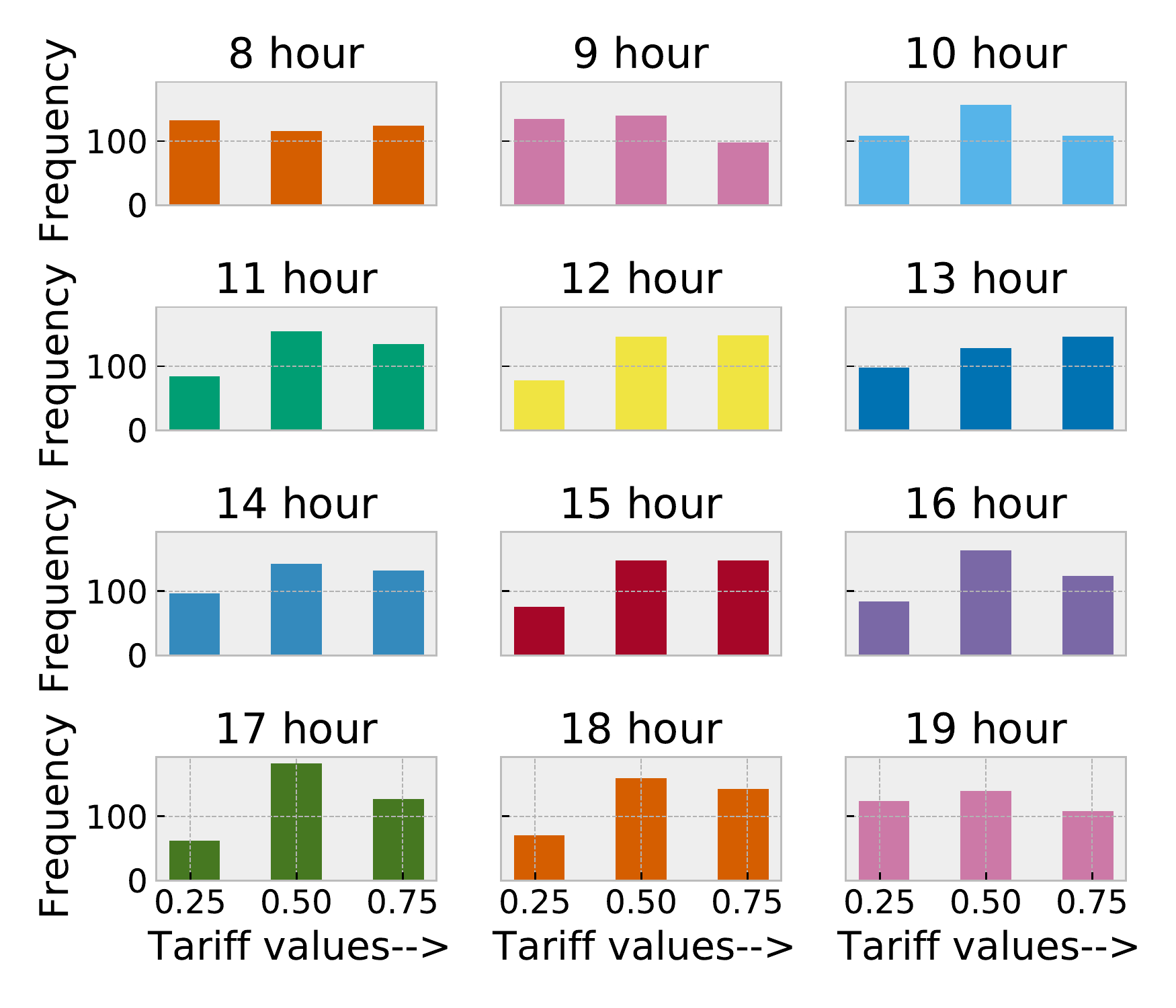}
        \caption{Hourly Tariff Distributions in OOD Profiles ($\mathcal{T}_{out}$).}
    \end{subfigure}
    \caption{(a) Logic for Consumption Data generation in Electricity Markets, and (b), (c) Hourly Tariff Rate Distributions depicting changing distribution across hours that poses generalization challenge. \label{fig:data}}
\end{figure*}
The aggregated consumption $e_{c,t}\in \mathbb{R}^+$ of a consumer $c$ at time $t$ has two components: i. \textit{Type-I consumption}: this is non-shiftable consumption corresponding to the appliances that have to be used at specific hours only and cannot be shifted to alternative hour, ii. \textit{Type-II consumption}: this is shiftable component of the consumption corresponding to appliances whose usage can be planned. Refer Figure \ref{fig:causaldiag} for more details.

Let $e_{c,1:t}$ denotes the time series of electricity consumption for consumer $c$ till time $t$.
We consider a consumer $c \in \mathcal{C}$, where $\mathcal{C}$ is the set of consumers with non-zero Type-II consumption, i.e. part of their load can be shifted in response to variations in tariff across hours.
Further, the $i$-th time-of-use (TOU) tariff profile is denoted as an ordered sequence or $H$-length time series of hourly tariffs $TOU^i=TOU^i_{1}\ldots TOU^i_H$, where $TOU^i_h$ ($h=1\ldots H$) denotes the tariff at hour $h$. In this work, we consider tariff profile with hourly rates over a day such that $H=24$, without loss of generality. 

Let $\textbf{f}_{c,1:t}$ denote all features (static or time-varying) for consumer $c$ at time $t$, including e.g. past consumption time series, type of consumer (household, office, etc.), and $\textbf{f}_t$ denote a vector of temporal features at timestamp $t$, e.g. hour of the day, day of the week, week of the month, month of the year, etc.
Note that $\textbf{f}_{c,1:t}$ refers to relevant features from entire history, but in practice, we consider a window of length $w$ over $t-w+1:t$ for deriving features at time $t$.

Further consider a tariff allocation policy function $\pi$ such that $TOU_{c,t+\tau} = \pi(\textbf{f}_{c,:t},\textbf{f}_{t+\tau},\hat{p}_{t+1:t+H})$, i.e. the tariff at a future time $t+\tau$ with $\tau=1\ldots H$ is decided based on consumer features at time $t$, the temporal features for time $t+\tau$, where $\hat{p}_{t+\tau}$ denotes the estimate of electricity price ${p}_{t+\tau}$ in the wholesale market at time $t+\tau$.
Without loss of generality, we consider the scenario where $t+1$ corresponds to first hour of the day, i.e. tariff profile for next day is decided using data till end of current day.

Consider historical time series data $\mathcal{D} = \{e_{c,1:t},TOU_{c,1:t}\}_{c\in\mathcal{C}}$, where the tariff time series are a result of sequence of tariff profile allocations over days such that any profile $TOU^i \in \mathcal{T}_{in}$ is chosen from a fixed set of profiles $\mathcal{T}_{in}$.

The goal for the broker is to allocate that tariff profile $TOU^i$ to a consumer that maximizes the gain $G^i_c$ over the next $H$ hours:
\begin{equation}
    G^i_c =  \sum_{t'=1}^H (TOU^i_{c,t+t'}-p_{t+t'})\times e_{c,t+t'}.
\label{eq:obj}\end{equation}

Importantly, the electricity consumption $e_{c,t+t'}$ at $t+t'$ hour is a function of the entire tariff profile on that day, as the consumer could choose to shift the shiftable part of the load from high tariff hours to low tariff hours by looking at the tariff profile allocated to the consumer at the beginning of the day. 

We consider the following two scenarios depending on the tariff profiles being considered for future allocations:
\newline
\textbf{IID Scenario}: when the profiles to be allocated to the consumers in future are from the same set of profiles $\mathcal{T}_{in}$ used historically, i.e. $TOU^i \in \mathcal{T}_{in}$.
\newline
\textbf{OOD Scenario}: when the tariff profiles to be allocated to the consumers in future belong to $\mathcal{T}_{all} = \mathcal{T}_{in} \cup \mathcal{T}_{out}$, where $\mathcal{T}_{out}$ is a new set of profiles not previously seen in $\mathcal{D}$, i.e. are out-of-distribution w.r.t. the training data, and not previously allocated to any consumer by the broker who wants to consider these new profiles to improve future gains, i.e. $TOU^i \in \mathcal{T}_{all}$.

\section{Related Work\label{sec:rw}}
Our work relates to two bodies of literature: i. demand response management in electricity markets and the related sub-problem of electricity consumption forecasting under exogenous variables using reinforcement learning and deep learning methods \cite{lu2019incentive,lu2018dynamic,yang2012game}, and ii. out-of-distribution (OOD) generalization \cite{hendrycks2021many,arjovsky2020out,krueger2021out,sun2019test}.

There have been lots of studies for (i), however, to the best of our knowledge, the problem of bias in historical data in terms of the tariff profiles has been largely overlooked.
We draw attention of the community working on (i) to the potential of OOD generalization by improving forecasts for previously unallocated tariffs by using the underlying structure of the problem in terms of the particular way in which consumers shift loads in response to changes in tariff. More specifically, we rely on the partial permutation equivariance property of the response to time series of tariffs. 

OOD detection and generalization is an emerging area of research, and aims at improving the robustness of models to previously unseen scenarios. 
Many of the recent approaches for (ii) rely on changes in the objective function or different training procedures.
For example, the approaches based on meta-learning \cite{hospedales2020meta} are not applicable as there is no notion of multiple tasks. We can consider each tariff profile as a task but then the forecasting can involve different profiles in input versus output. In this work, we focus on using inductive biases in the form of the neural network architecture to improve OOD generalization.
There is enough evidence to support the improvement in generalization abilities of neural networks by using the structure of the problem to introduce suitable inductive biases in the learning process. The most commonly used inductive bias is in the design of the neural network architecture motivated by the structure of the problem. Recent examples of this include using graph neural networks \cite{wang2018nervenet,narwariya2020graph} and modular networks \cite{andreas2016neural}. Recently, using structural biases in deep neural networks motivated by the nature of bias and the structure of the problem have been successfully evaluated for time series forecasting \cite{bansal2021systematic}.
Data-dependent priors have been recently proposes in \cite{liu2021statistical}. 
However, to the best of our knowledge, using consumer behavior properties for electricity time series forecasting under out-of-distribution exogenous variables to guide the design of neural network architecture has not been considered so far in literature.

\section{The Learning Problem}

We consider a 2-step approximate solution to maximize the gain (Eqn. \ref{eq:obj}):
\newline
\textbf{Step-1}: For each consumer, forecast/estimate the consumption under each potential tariff profile allocation. 
Given features $\mathbf{f}_{c,1:t}$ (including $e_{c,1:t}$), history of allocated tariffs $TOU_{c,1:t}$, and values of potential future tariff $TOU_{c,t+1:t+H}$, the goal is to estimate $e_{c,t+1:t+H}$.
This can be seen as \textbf{a multi-step time series forecasting problem with exogenous variables}. We provide details of our proposed approach for this in the next section.
\newline
\textbf{Step-2}: Compute the profit using 
\begin{equation}
 \hat{G}_c^i = \sum_{t'=1}^H (TOU^i_{c,t+t'}-\hat{p}_{t+t'})\times \hat{e}_{c,t+t'}
\end{equation}
for each tariff in $TOU^i \in \mathcal{T}_{all}$ for OOD scenario ($\mathcal{T}_{in}$ for IID scenario).   
Allocate the tariff profile to consumer $c$ which results in maximum $\hat{G}^i_c$.
Note that in practice the future wholesale rates $p_{t+t'}$ ($t'=1\ldots T$) are also not known and might need to be estimated. In this work, we assume that $p_{t+t'}$s are known in advance or estimable accurately and focus on estimating $\hat{e}_{c,t+t'}$s which are the only terms controllable via $TOU_{c,t+t'}$s.

In summary, the tariff profile allocation policy corresponds to estimating the gain for each tariff profile for a consumer, and then allocating the profile with maximum estimated gain.
We use a deep neural network based architecture as the function approximator that estimates  $\mathbb{E}[e_{c,t+t'}|TOU_{c,t+1:t+T}]$ from the data.

\begin{figure*}[h!]
    \centering
    \includegraphics[width=16cm, trim={2.0cm, 10.5cm, 2.0cm, 1.0cm}]{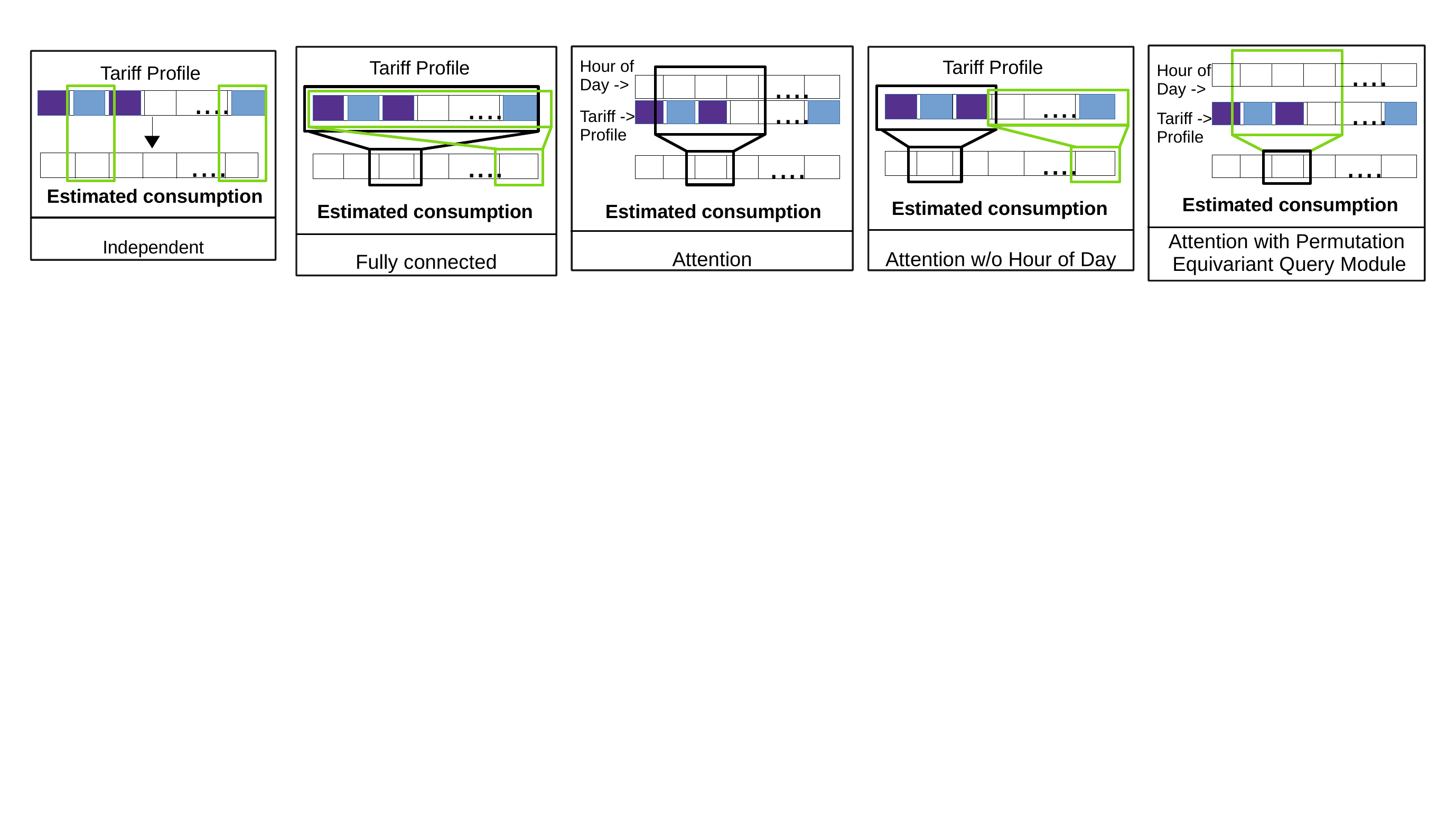}
    \caption{ How different methods process the sequence of tariff rates.\label{fig:framework}}
    \label{fig:Different_Methods}
\end{figure*}

\subsection{Biased and Scarce Data}
The OOD scenario is challenging as there is no historical data for the profiles in $\mathcal{T}_{out}$. 
More concretely, we consider three possible values of tariff at any time $t$: low (0.2), medium (0.5), and high (0.8). Therefore, there are $3^H$ unique profiles possible. For $H=24$, there can be $\approx 3\times 10^{11}$ profiles possible. However, in practice, the number of allocated profiles would be significantly smaller than this. In this work, we consider $|\mathcal{T}_{in}| \in \{2,5,8,10,12,15,20,30,35\}$ which is a range of values encountered for $|\mathcal{T}_{in}|$ in practice. This poses serious OOD generalization challenge in estimating $e_{c,t+1:t+T}$ for previously unseen profiles $TOU^i_{t+1:t+T} \in \mathcal{T}_{out}$. 

We note that one peculiar type of bias that manifests in practice is the \textbf{temporal bias}: at any hour $h$ of the day, certain values of tariff are more common than others. We explain this further using a practical scenario as depicted in Figure \ref{fig:data}: In practice, it is common to use the following heuristic for tariff profile allocation: Keep most expensive tariff rates during peak demand periods, least expensive tariff rates during non-peak hours, and slightly cheaper (medium) rates, typically between peak and off-peak periods. Every tariff profile is curated on the basis of average aggregated consumption of each customer. High tariff is allocated when the aggregated consumption is high, and for rest of hours low/mid tariff are allocated. The distribution of tariff rates over hours would depend on the distribution of peak consumption across customers (refer Figure \ref{fig:data}c). Furthermore, there is confounding bias \cite{pearl2016causal} with latent consumer attributes affecting i. past aggregated consumption which in turn affects the treatment (tariff profile allocation), and ii. the outcome (electricity consumption) in $\mathcal{D}$ both can depend on the consumer features (refer Figure \ref{fig:causaldiag}). We leave handling of confounding bias for future work, and focus on handling temporal bias in this work.

We empirically show that temporal bias poses generalization challenge for vanilla feedforward neural networks, and propose an attention-based architecture to deal with the same, in the next section.

\begin{figure*}[h!]
    \centering
    \includegraphics[width=16cm, trim={2.0cm, 6.5cm, 0.2cm, 2.0cm}]{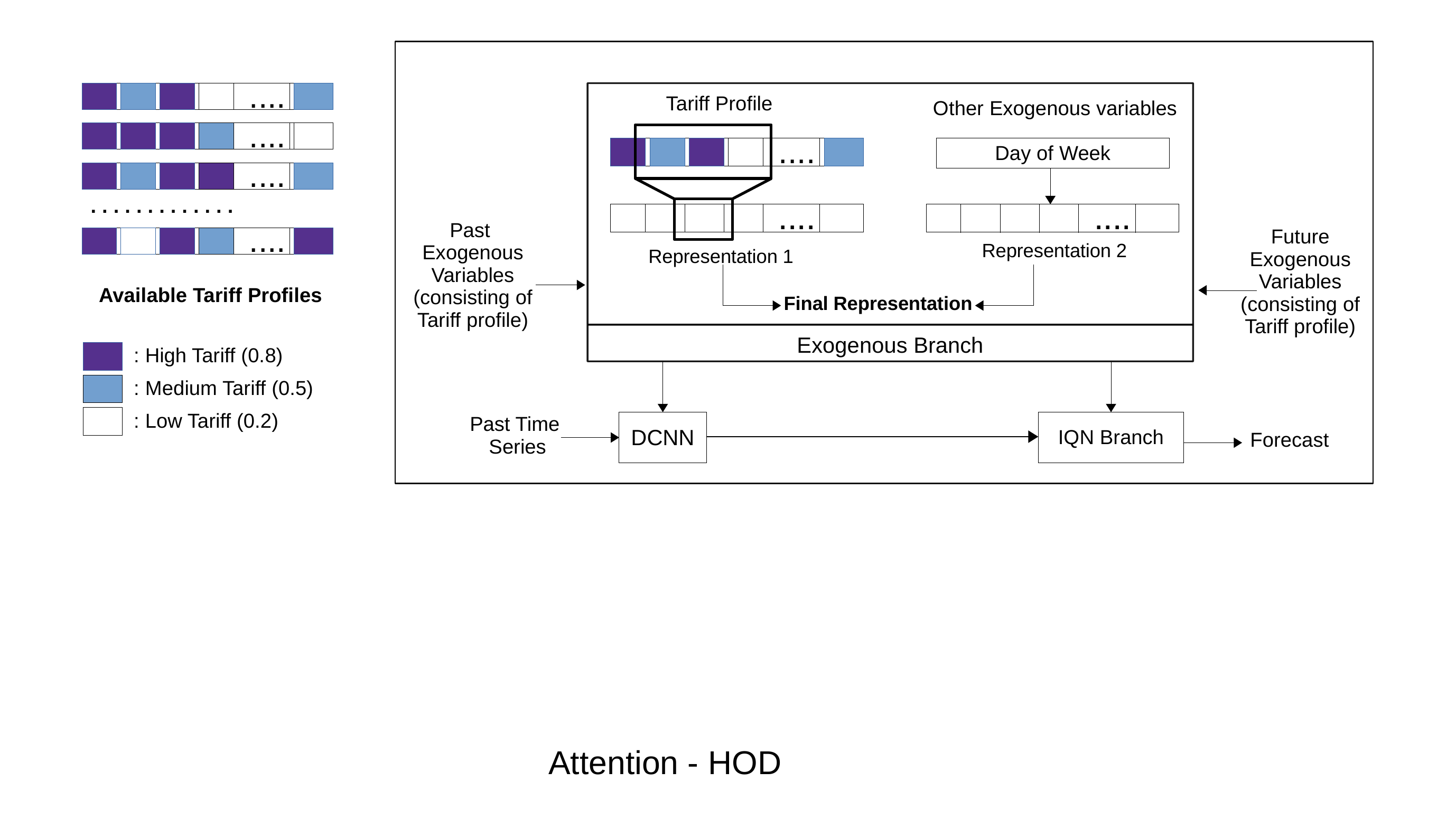}
    \caption{ Flow diagram of ``Attention w/o Hour of Day'' approach. Left part of the figure indicates the variability in the tariff profiles and also some tariffs are more frequent in tariff profiles. 
Right part of the figure indicates flow of the inputs through the network and how the information of tariffs is consumed by the proposed approach.\label{fig:framework}}
\end{figure*}

\begin{figure*}[h!]
    \centering
    \includegraphics[width=16cm, trim={0.8cm, 6.0cm, 0.2cm, 0.2cm}]{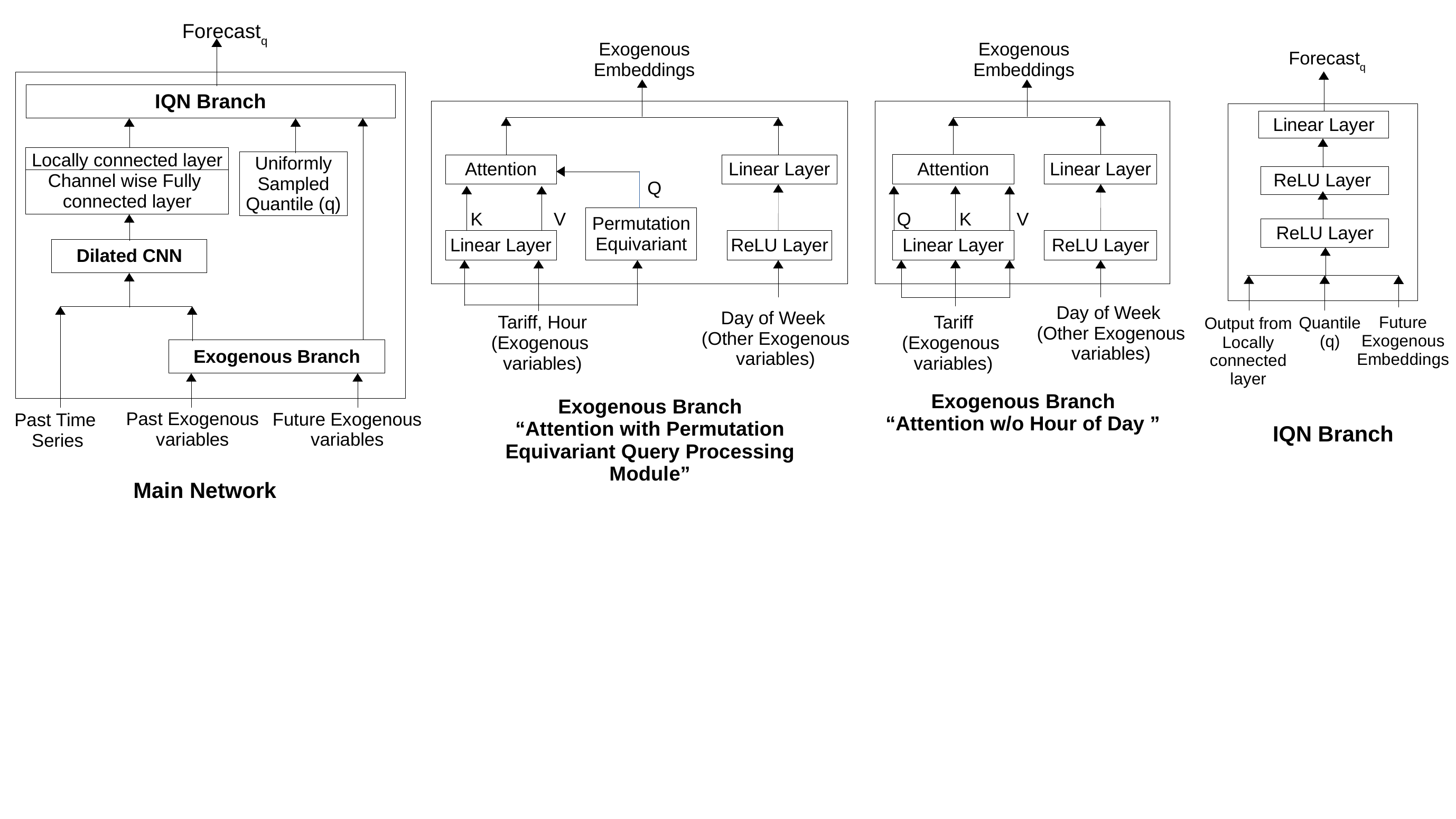}
    \caption{Architectures contrasting ``Attention w/o Hour of Day" and ``Attention with Permutation Equivariant Query Processing Module" approaches \label{fig:archdetails}}
\end{figure*}
\subsection{How consumers respond to Tariffs?}
Consider the following toy example with $H=6$ where there is only one tariff profiles in $\mathcal{T}_{in}$ given by \textit{\{HHMMLL\}}, i.e. tariff rate is high (H) for first two hours, medium (M) for next two hours, and low (L) for last two hours. Further assume that consumer has a certain Type-II load during the 1st hour. After looking at this tariff profile, consumer responds by shifting the load from 1st (high tariff) hour to 5th (low tariff) hour.
Now, consider a tariff profile in $\mathcal{T}_{out}$ as \textit{\{HHLLMM\}}. Clearly, this profile is different from the profile in $\mathcal{T}_{in}$ as the sequence of highs and lows over the hours is different.
However, importantly, the underlying decision-making behavior of the consumer remains the same, i.e. shift the Type-II load from high tariff hour (1st hour in this case) to low tariff hour (3rd hour instead of 5th hour in this case).
Therefore, it is still possible to forecast the behavior of the user for this OOD profile.
In this work, we intend to leverage this aspect of the consumer's decision-making process that stays same irrespective of the IID-vs-OOD profiles.

Further, consider five ways to process the sequence of tariff rates (Figure \ref{fig:Different_Methods}):
\begin{itemize}
    \item \textbf{Independent processing}: here, the tariff at each hour is processed independently \cite{salinas2020deepar,liu2020deepssm} and used to estimate the consumption at that hour. Of course, since consumer's decision-making is based on comparison of tariff rates across hours, such a processing of tariff profiles will not be effective.
    \item \textbf{All considered together or fully connected}: here, tariffs at all hours (the entire tariff profile) is processed simultaneously, e.g. through a fully connected layer in a feedforward neural network. We argue that such processing of tariff profiles will be able to effectively learn a good function approximator for the profiles in $\mathcal{T}_{in}$. However, it will be highly biased to the profiles in $\mathcal{T}_{in}$ since it does not effectively learn the way consumers are processing the tariff rates for shifting the loads. This leads to biased tariff profile processing modules due to the temporal bias in the historical profiles, as discussed above.
    
    \item \textbf{Focusing on relevant information or Attention}: here, the tariffs rates on a day are considered as tokens and hours of a day are used as a positional information. This information is processed through a self-attention layer. We argue that such processing of tariff profiles will mimic the logic of how consumers respond to a tariff profile. However, it will be biased towards the profiles in $\mathcal{T}_{in}$ since the tariffs and hour of the day are correlated (due to temporal bias in the historical tariff profiles).
    
    \item \textbf{Permutation Equivariance}: as discussed earlier, permutation equivariance is an important aspect of the consumer decision-making logic. To mimic the same in the processing of tariffs by the neural networks, we expect that if trained on one of the tariff sequences say, \textit{HHMMLL} in the earlier example) it should perform equally well on other sequence (i.e. \textit{HHLLMM}). In other words, processing of tariffs by neural networks should be Permutation Equivariant. We propose two ways to achieve approximate permutation equivariance:
    \begin{itemize}
        \item \textbf{Attention w/o Hour of Day (Att.-HOD) }: As explained above, the standard self-attention method can mimic the logic of how consumers respond to tariffs but due to temporal bias in the data attention method does not generalize well to $\mathcal{T}_{out}$. We propose a simple variant that does not take HOD as input in the self-attention module to obtain the permutation equivariance property.
        \item \textbf{Attention with Permutation Equivariant Query Processing Module (Att.+PE)}: here, the tariff rates on a day are considered as a set, and processed in such a way that ordering of the tariff rates does not matter, i.e. the processing is permutation equivariant \cite{zaheer2017deep,lee2019set}.
    \end{itemize}

\end{itemize}

In the next section, we explain how we achieve permutation equivariance while forecasting the consumption given consumer's consumption history, sequence of past tariff profiles, and a future tariff profile.

\section{Forecasting Architecture  \label{sec:approach}}

Consider the consumption history of a consumer along with past allocated tariffs to be a time series of vectors $\mathbf{f}_{1:t}$ including dimensions for past aggregate consumption and past allocated tariff rates $\{e_{1:t},TOU_{1:t}\}$, and the candidate tariff profile for the next $H$ hours to be $TOU_{t+1:t+H}$.
The goal is to estimate $e_{t+1:t+H}$ while ensuring permutation equivariance in processing $TOU_{1:t+H}$ in the sense of \cite{lee2019set}, e.g. if the output of processing $\{TOU_1,TOU_2,TOU_3\}$ is $\{\mathbf{o}_1,\mathbf{o}_2,\mathbf{o}_3\}$, then the output of processing a permutation of the input, say $\{TOU_2,TOU_1,TOU_3\}$ is given by the permutation $\{\mathbf{o}_2,\mathbf{o}_1,\mathbf{o}_3\}$ of the original output. 

To achieve the above-stated goal, we consider the following modularized neural network architecture as depicted in Figures \ref{fig:framework} and \ref{fig:archdetails}:
\begin{itemize}
    \item \textbf{Dilated Convolutional Neural Networks (DCNN)\footnote{Since we have large input time series (t=168 in our case), we consider 1D-Convolution Neural Networks for computational efficiency instead of Recurrent Neural Networks based architecture such as LSTMs \cite{hochreiter1997long}.}} branch for processing of past consumption time series.
    \item \textbf{Exogenous} branch:
    This branch consists of \textbf{Attention with Permutation Equivariant Query Processing Module (Att.+PE)} branch for processing of tariff rates, and other modules for processing of features like hour of day, day of week, etc.
    \item \textbf{Implicit Quantile Network (IQN)} branch for generating the quantile estimates for future consumption. 
\end{itemize}

Next, we provide details of the exogenous branch which is the key novel component of the proposed approach and helps to mitigate temporal bias.

To achieve permutation equivariance and handle temporal bias, we consider processing the tariff rates $TOU_{t+1:t+H}$ (same processing is done for past tariffs as well) via an attention mechanism where a part of the processing is done independently for tariff at each time step $t+t'$ ($t'=1\ldots H$) while still taking into account the global information $TOU_{t+1:t+H}$ in order to mimic the behavior of the consumer as explained in the previous section.

More specifically, we consider key $K$ and value $V$ for the attention mechanism to be dependent on a single time step $t+t'$, while the query $Q$ depends on the entire tariff profile $TOU_{t+1:t+H}$ for the day. In other words, $K_{t+t'}=f_K(TOU_{t+t'}, t+t', \theta_K)$, $V_{t+t'}=f_V(TOU_{t+t'}, t+t', \theta_V)$, and $Q_{t+t'}=f_Q(TOU_{t+1:t+H}, \theta_Q)$.
Subsequently, the output for the part of the exogenous branch processing the tariffs at time $t+t'$ is given by 
\begin{equation}
 \texttt{Att}(Q_{t+t'},K_{t+t'},V_{t+t'})= \texttt{softmax}(\frac{Q_{t+t'}K_{t+t'}^T}{\sqrt{d}})V_{t+t'},
\end{equation}
where $d$ is the dimension of Q, K, and V.
While the $f_K$ and $f_V$ are implemented as simple linear layers, $f_Q$ is implemented as a permutation equivariant network as follows:
\begin{equation}
    f(x) = \sigma(x\mathbf{\Lambda}-\bold{1}\texttt{maxpool}(x)\mathbf{\Gamma})
\end{equation}
where $x = \texttt{ReLU}(TOU_{t+1:t+H},\theta_{TOU}) \in \mathbb{R}^{H\times d}$ and $\theta$ shared across timesteps $t+1\ldots t+H$, $\mathbf{\Lambda}, \mathbf{\Gamma}\in \mathbb{R}^{dxd'}$, matrix of ones $\bold{1} \in 1^{H\times H}$, $\texttt{maxpool}$ is taken along columns implying that the resulting value for any timestep contains information from all timesteps and is independent of a particular timestep. In this work, we use $d=10$, $d'=20$.

\textbf{Objective function}:
We use quantile loss for training the DCNN model given by:
\begin{equation}\label{eq:loss}
\mathcal{L}_{quantile} = \frac{1}{b\times n}{\sum_{i=1}^{b}{\sum_{q=q_1}^{q_n} {max (q\times e^i,(q-1)\times e^i)}}},
\end{equation}
where $e^i=y^i-\hat{y}^i$ indicates the error of the forecasted consumption $\hat{y}^i$ w.r.t. ground-truth consumption $y^i$ of $i$-th window instance, $b$ is the batch size and $n$ is the number of quantiles used for training.

\section{Experimental Evaluation}
\begin{figure*}[h!]
   \captionsetup[subfigure]{font=small,labelfont=small, justification=centering}
    \centering
    \begin{subfigure}{0.43\textwidth}
        \centering
        \includegraphics[scale=0.45,trim={0.23cm 0.3cm 0.23cm 0.24cm},clip]{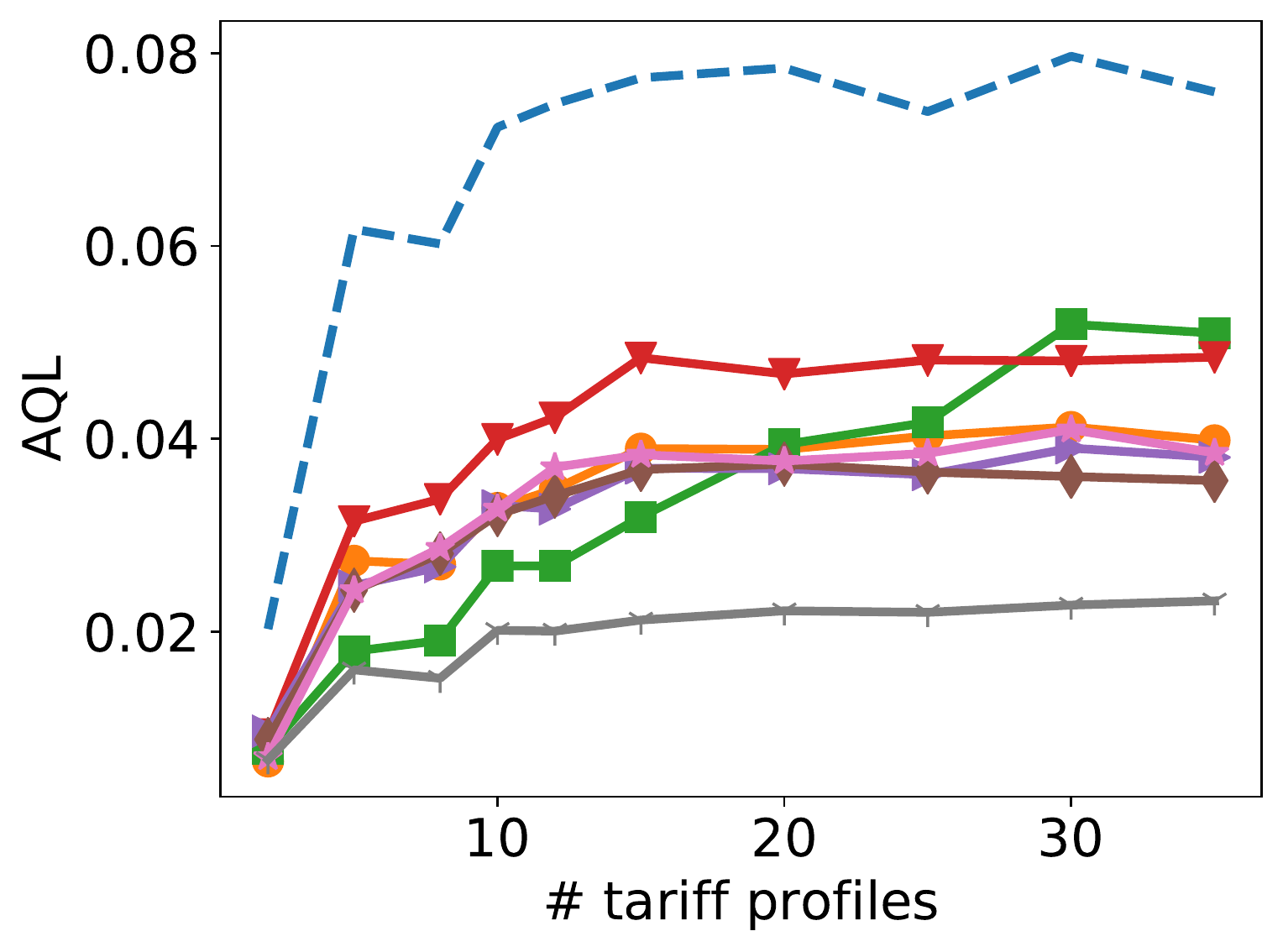} 
        \caption{IID Scenario}
    \end{subfigure}
    \begin{subfigure}{0.5\textwidth}
        \centering
        \includegraphics[scale=0.45,trim={0.87cm 0.3cm 0.2cm 0.26cm},clip]{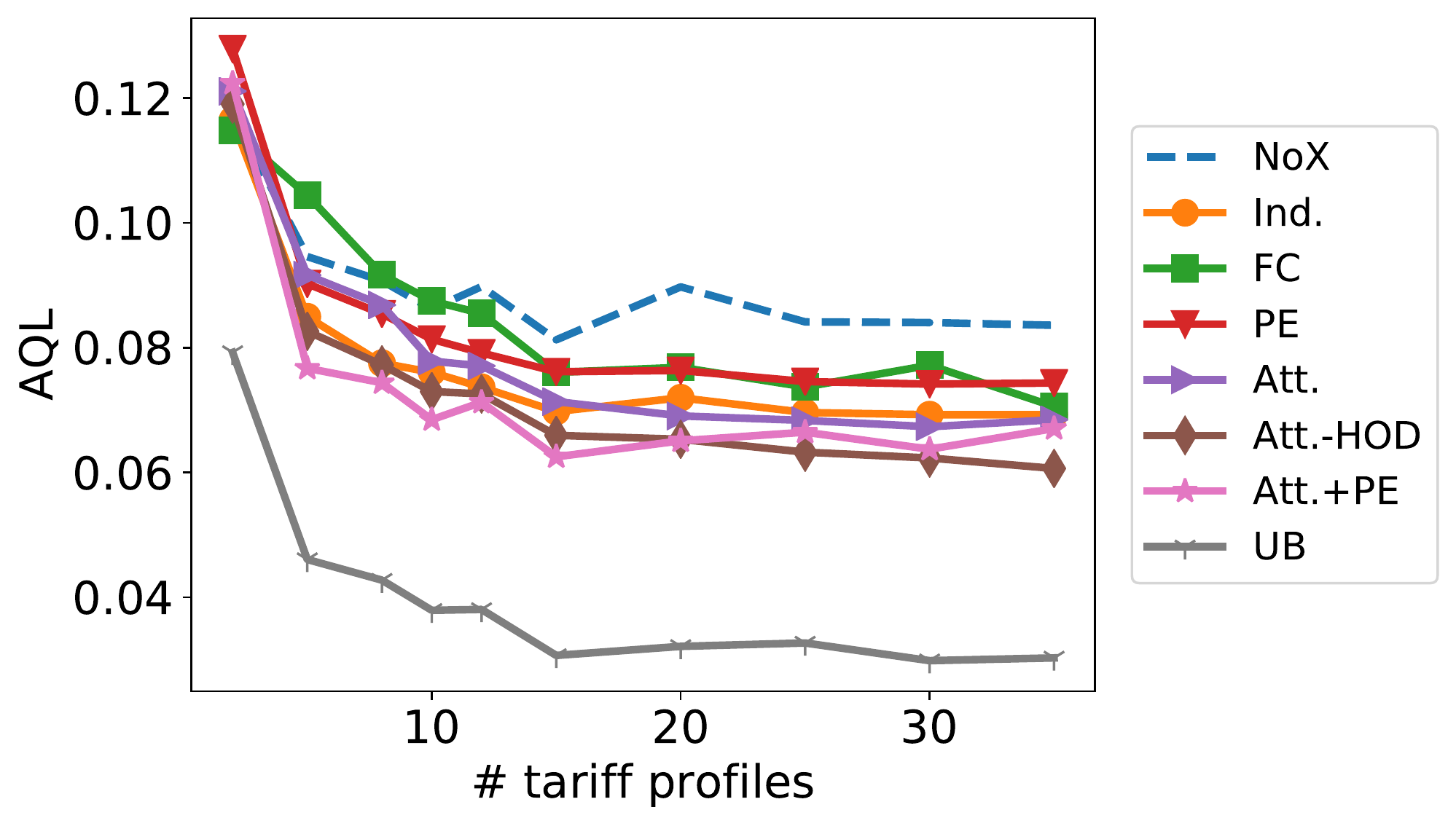} 
        \caption{OOD Scenario}
    \end{subfigure}
    \caption{Forecasting performance Comparison of different approaches (in terms of Average Quantile Loss) \label{fig:Res_forecast}}
\end{figure*}
\begin{figure*}[h!]
   \captionsetup[subfigure]{font=small,labelfont=small, justification=centering}
    \centering
   
    \begin{subfigure}{0.45\textwidth}
        \centering
        \includegraphics[scale=0.45,trim={0.23cm 0.5cm 0.4cm 0.2cm},clip]{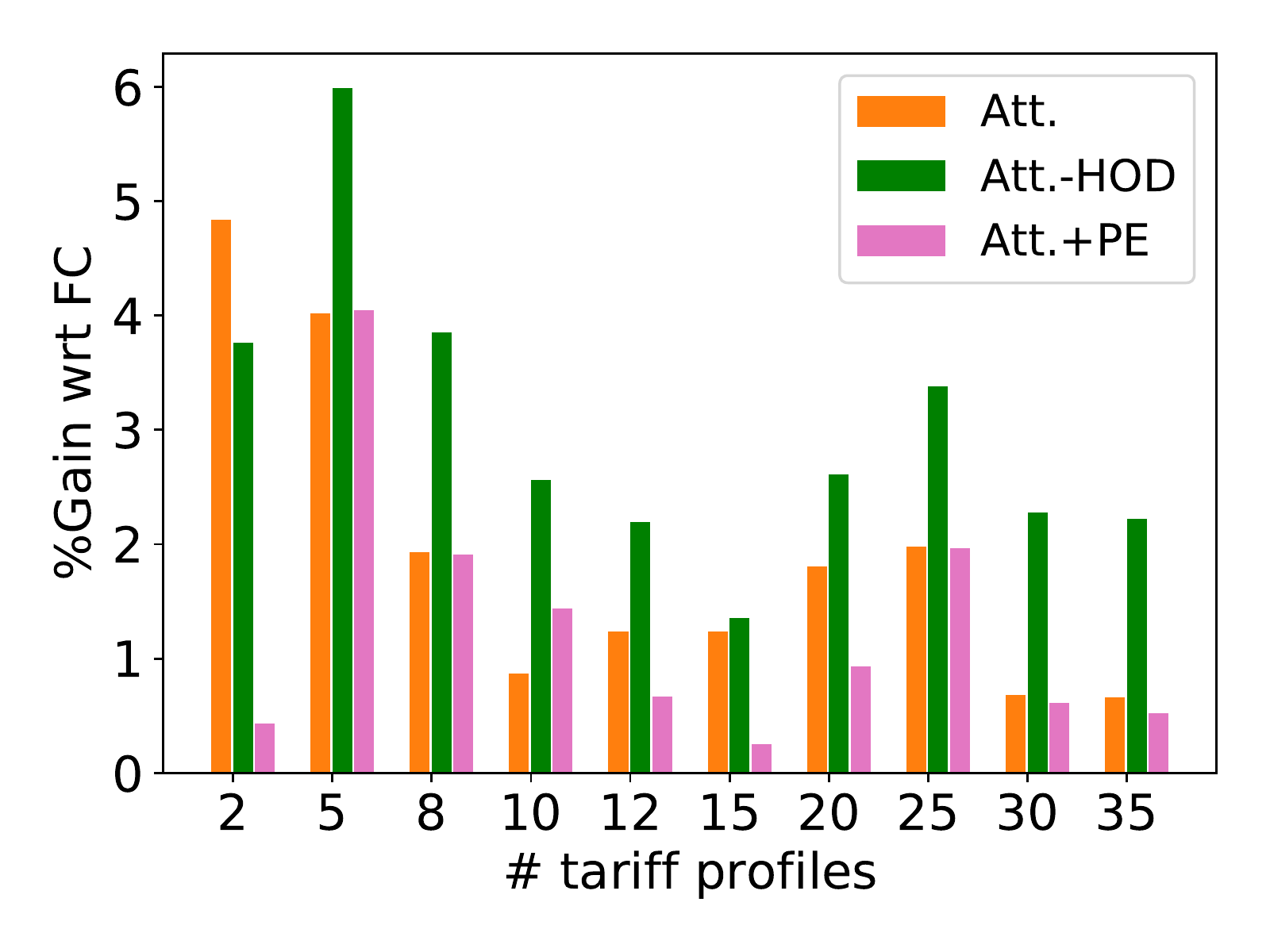}
        \caption{Option-1}
    \end{subfigure}
    \begin{subfigure}{0.45\textwidth}
        \centering
        \includegraphics[scale=0.45,trim={1.3cm 0.5cm 0.4cm 0.2cm},clip]{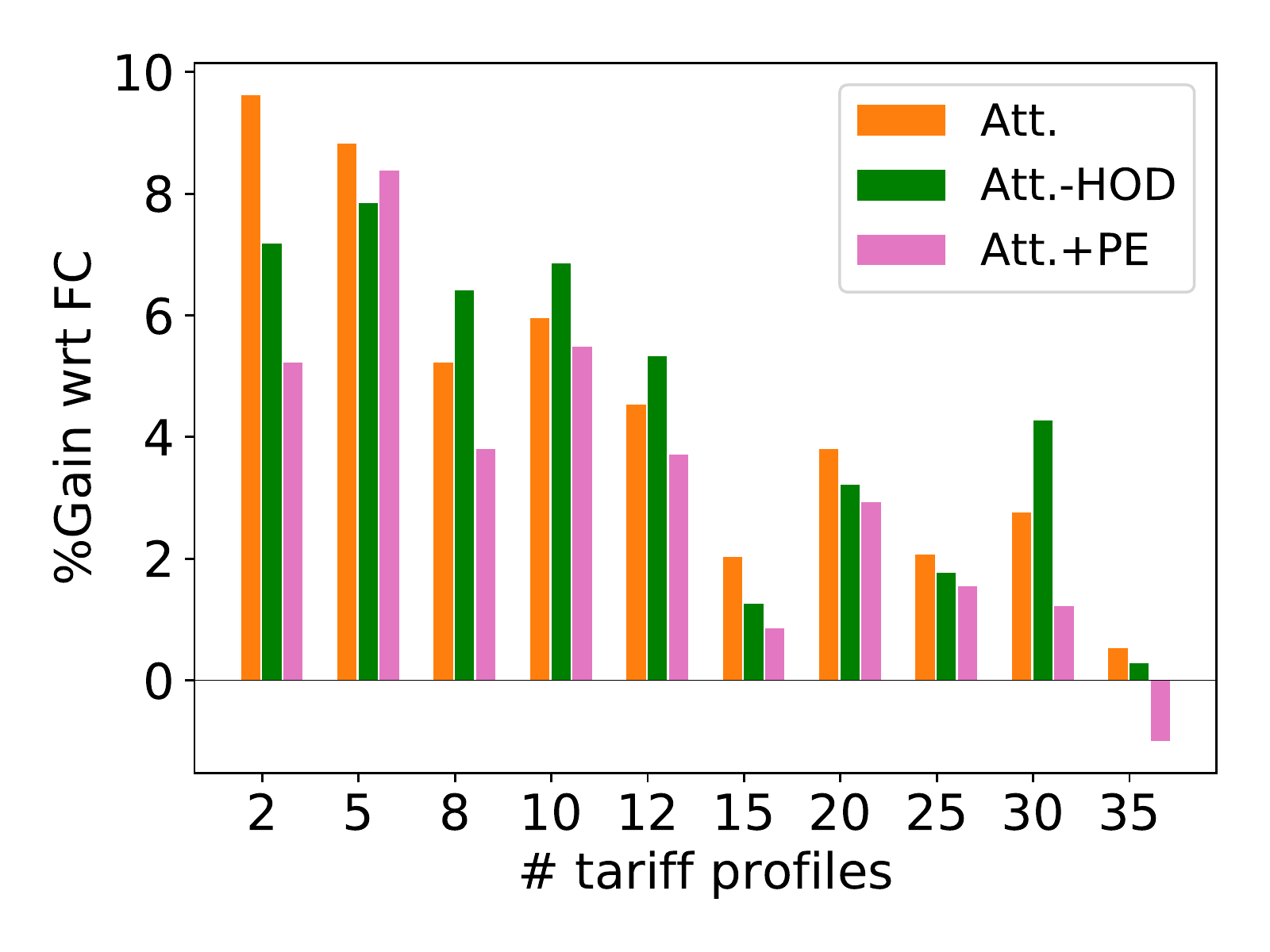}
        \caption{Option-2}
    \end{subfigure}
    \caption{ \%gains of the proposed Att.+PE, Att.-HOD and Att. approaches over the vanilla FC approach. \label{fig:SC1_cost}}
\end{figure*}
The goal is to evaluate the efficacy of the proposed approach to deal with OOD scenarios. For this, we compare the proposed approach with various baselines in the IID as well as OOD settings. 
We use the simulated data from a high-fidelity and popular PowerTAC\footnote{https://powertac.org/} \cite{ketter2013power} simulator that uses complex state-of-the-art user-behavior models and real world weather data to simulate the complex dynamics of a smart grid system. 

We consider `Office Complex Controllable type' consumers where consumers' daily behavior depends on factors such as number of sub-customers, number of appliances, weather information, hour of day, month, day of week, etc. The various values these factors can take across consumers is given in Table \ref{tab:data}.
\begin{table}[!h]
    \centering
    \caption{Dataset details}
    \scalebox{0.8}{
    \begin{tabular}{|c|c|c|}
    \hline
         S.N. &  Properties of consumers& Value(s) \\\hline
        1 &Number of consumers & 12 \\
        2 &Number of sub-consumers & 3, 5 \\
        3 &Working days & 3, 4 \\
        4 & Work Start hour  & \{8, 9, 10\} (+/-) 1 hours \\
        5 & Break Start hour & \{13, 14\} (+/-) 1 hours \\
        6 &Work duration & 8 (+/-) 1 hours \\
        7 &Shiftable consumption( in KW) & 600, 2400 \\
        8 &Total data duration (in months)& 6 \\\hline
    \end{tabular}}
    
    \label{tab:data}
\end{table}

To obtain train, validation and test split, we divide total data of 6 months in 4, 1 and 1 month, respectively.
The time series of hourly data for each consumer is divided into windows of length t = 168 (corresponding to 7 days) with window-shift of 24 to forecast one day-head consumption, i.e. output window size is 24.
We consider varying number of tariff profiles in historical data, i.e. $|\mathcal{T}_{in}| \in \{2,5,8,10,12,15,20,25,30,35\}$, and an additional set of $|\mathcal{T}_{out}|=40$ profiles. As the number of profiles $|\mathcal{T}_{in}|$ in the training set increases, we expect the bias in the training data to reduce.

\subsection{Baselines Considered}
For comparison, we consider following approaches all using DCNN as the core time series processing module:
\begin{itemize}
\item \textbf{No future exogenous variable (NoX)} is the simple univariate time series forecasting approach which uses only history of aggregated consumption without any additional future information.
This can be considered as a lower bound in the sense that the network does not have access to any future tariff rates to estimate where a consumer will shift the load.

\item \textbf{Independent tariff-based method (Ind.)} is an approach that treats each tariff rate independently, and uses the tariff at time $t+t'$ to estimate the aggregated consumption at that time. Importantly, this approach has no means to capture comparison of the tariff rates in order to figure out whether the tariff at time $t+t'$ is high or low in comparison to another timestep. 

\item \textbf{Fully-Connected Approach (FC)} utilizes the information of all timesteps to estimate the aggregated consumption at each timestep. As explained previously, we expect such an approach to perform well in the IID scenario but struggle in the OOD scenario where new profiles are included. 

\item \textbf{Permutation Equivariant (PE)} method uses only the permutation equivariance idea from our approach and ignores the attention mechanism. This method can be thought of as an ablation over our approach. 

\item \textbf{Attention (Att.)}: This is another ablation over our approach which uses standard attention module for processing the tariffs along with hour of the day information without any permutation equivariance property. 

\item \textbf{Upper Bound (UB)}: This is an oracle approach that assumes knowledge about the hours at which the consumer is going to shift the load. In this, a binary value indicating whether the shiftable load will be shifted to this hour or not is passed as an additional feature to the exogenous branch of the Att.+PE network.
\end{itemize}

\begin{figure*}[h!]
   \captionsetup[subfigure]{font=small,labelfont=small, justification=centering}
    \centering
    \includegraphics[width=14cm, trim={0.1cm, 0.5cm, 0.8cm, 0.9cm}]{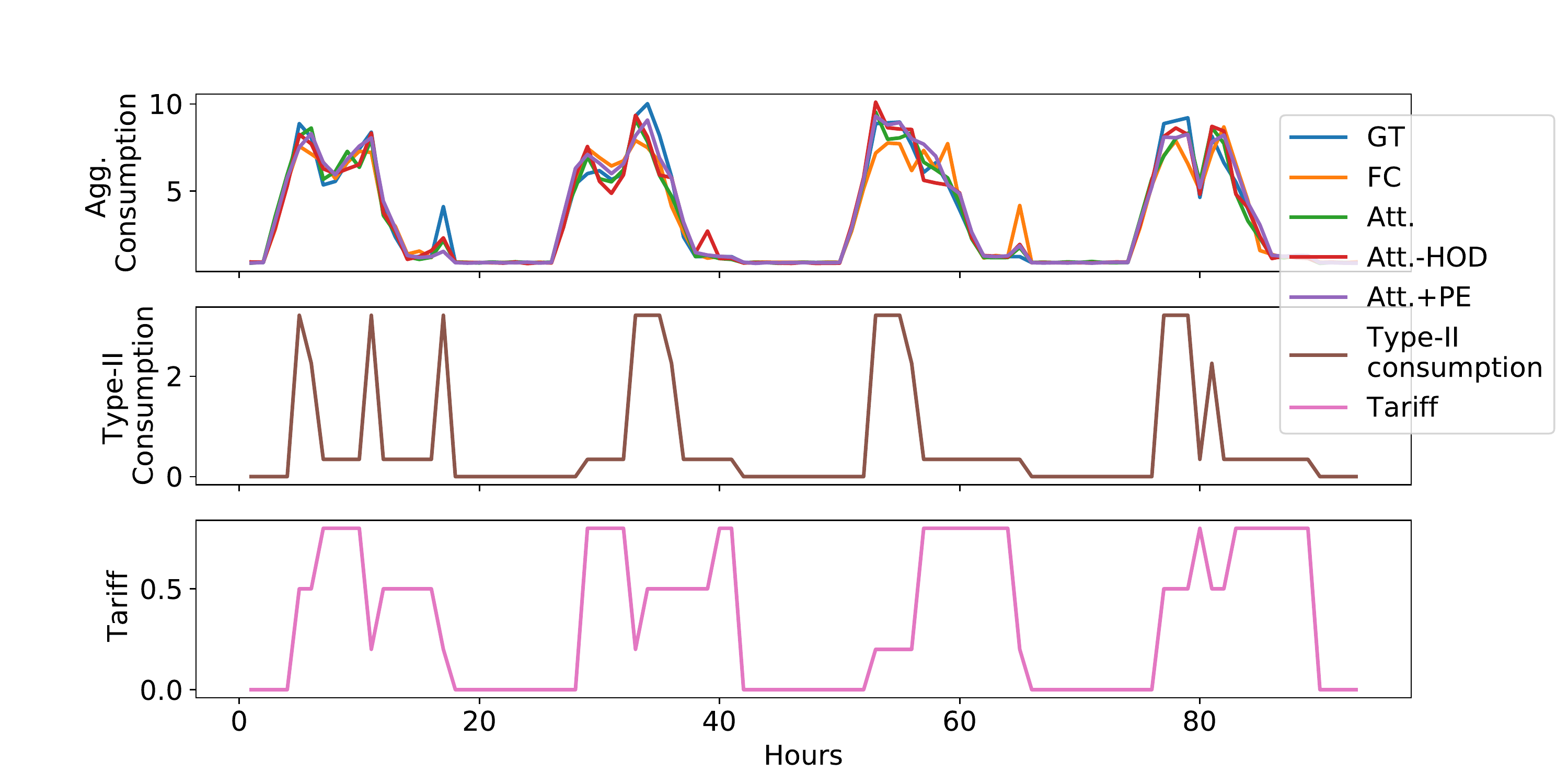}
    \caption{Sample results comparing the proposed approaches Att.-HOD and Att.+PE with FC on an OOD tariff profile. Here, GT: Ground Truth time series. FC struggles to capture the subtle changes in consumption due to shifting of load, while both Att.-HOD and Att.+PE are able to forecast better. \label{fig:forecast}}
\end{figure*}

\begin{figure*}[h!]
   \captionsetup[subfigure]{font=small,labelfont=small, justification=centering}
    \centering
    \includegraphics[width=14cm, trim={0.1cm, 0.5cm, 0.8cm, 0.9cm}]{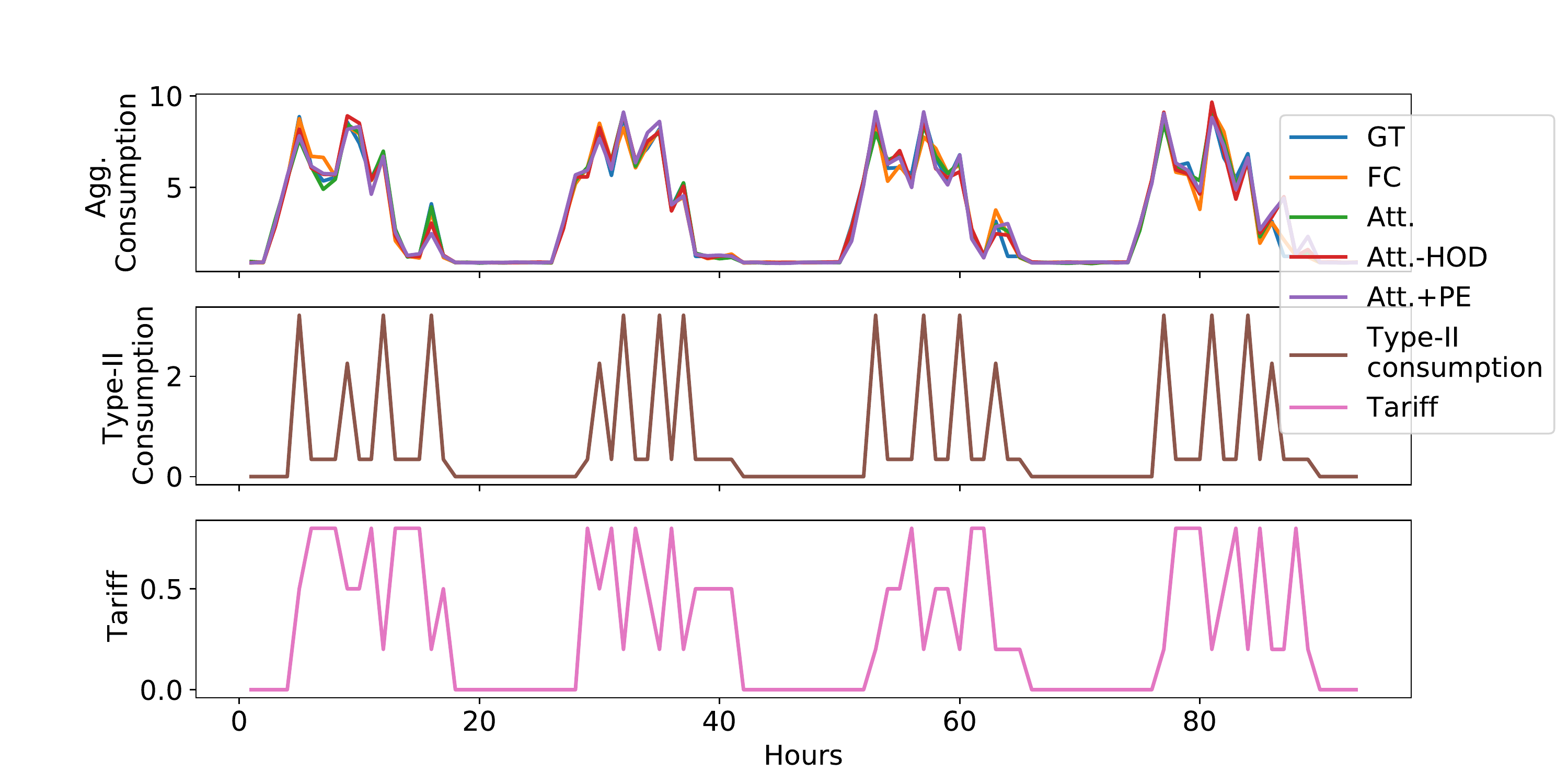}
    \caption{Sample results comparing the proposed approaches Att.-HOD and Att.+PE with FC on an IID tariff profile. Here, GT: Ground Truth time series. In IID scenario, all proposed attention-based approaches and baselines perform well. \label{fig:iidforecast}}
\end{figure*}
\subsection{Hyperparameters Used}
We use z-normalized consumption time series. DCNN has 3 layers with each layer having 16 convolutional filters of length 2, and dilation rate 1, 2 and 4, respectively. We use batch normalization and L2 filter regularizer ($\lambda$=0.001) for regularization purposes. and ReLU layers applied on each CNN layer. 
The output of DCNN layer is processed by channel-wise fully connected layer, which has 24 hidden units (equal to the output window size) i.e. 24, followed by locally connected layer with 10 filters which applied at each time-step independently (filter size=1).

To obtain categorical features (hour of day, day of week, month of year) embeddings and tariff rate embeddings, we use separate feed-forward network with ReLU layer followed by linear layer, having 5 hidden units and 10 hidden units respectively. Similarly, we use 10 hidden units for each feed-forward network $f_{Q}$, $f_K$, $f_V$. 
Finally, the output layer is a small feed-forward network that has 2 layer followed by linear layer having 40, 10, and 1 hidden units, respectively.
We use batch size of 16, number of epochs 200 and Adam optimizer with fixed learning rate of  0.0001 for training the neural network. 
During training, quantiles are sampled from uniform distribution while during validation and testing, we use 3 quantiles 0.1, 0.5 and 0.9. All hyperparameters were obtained via grid search based on validation quantile loss on the IID set. 

\subsection{Results and Observations}
We make following key observations from the results in Figure \ref{fig:Res_forecast} and Figure \ref{fig:SC1_cost}:
\begin{itemize}
    \item Observations from forecasting results as shown in Figure \ref{fig:Res_forecast}:
    \begin{itemize}
        \item In the IID scenario, the average quantile loss (AQL) for all approaches increases with increasing number of tariff profiles as the complexity of the dataset increases. FC approach performs better than other approaches for $|\mathcal{T}_{in}| \leq 15$, indicating higher expressivity of the FC approach to fit to a smaller number of IID profiles, indicating potential overfitting.

        \item On the other hand, for the OOD scenario, the performance of all approaches improves with increasing number of IID profiles which is expected as more IID profiles implies less bias and better generalization to OOD profiles as well. Interestingly, FC approach which was the best approach for the IID profiles for  $|\mathcal{T}_{in}| \leq 12$, is the worst approach (except the lower bound NoX) in the OOD setting,    because it uses a fully connected layer to process the tariffs of the day and due to temporal bias in the data, the weights of fully connected layer will try to overfit on $|\mathcal{T}_{in}|$ and thus not generalize to OOD profiles$|\mathcal{T}_{out}|$.

        On the other hand, our proposed approaches Att.+PE and Att.-HOD are consistently better than FC for all values of $\mathcal{T}_{in}$, which shows that FC struggles with the temporal bias in the historical data. 
        We also analyze that Att.-HOD as well as Att.+PE are also consistently better than Att. for all values of $\mathcal{T}_{in}$, which shows that permutation equivariant way of handling tariff profiles provide better generalization on OOD profiles.

    \end{itemize}
    
     \item We further analyze whether the gains of Att.+PE and Att.-HOD over other methods on the OOD scenario translate into more profitable tariff profile allocation for the retailer, we compare the gain G of Att.+PE, Att.-HOD and Att. in comparison to FC. We consider two kinds of profiles for wholesale prices $p$, one with two values (0.2 and 0.8, referred to as Option-1) and one with three values (0.2, 0.5, and 0.8, referred to as Option-2). 
     \begin{itemize}
         \item \textbf{Comparison with FC}: 
         We observe that all attention-based proposed approaches Att., Att.-HOD and Att.+PE depict significant positive gains over FC. We also observe that Att., Att.-HOD and Att.+PE approaches have higher positive gain in less IID tariff profiles scenarios $|\mathcal{T}_{in}| \leq 12$ (except $|\mathcal{T}_{in}| = 2$ where data is too less to claim any generalization), and the gains tend to diminish as $|\mathcal{T}_{in}|$ increases.
          
         \item As expected, we note that it is not important that the gains in forecasting translate directly into monetary profits, as the optimization objective involves other terms such as wholesale costs $p$. Therefore, the best approach on forecasting (Att.+PE) in OOD scenario is not necessarily the best approach in terms of profit always.
         
         \item \textbf{Comparison with Att.}: For Option-1, Att.-HOD has significantly better gains than Att. for all values of $\mathcal{T}_{in}$ except $|\mathcal{T}_{in}| = 2$, which shows that the permutation equivariant way of handling tariff profiles is helpful. For Option-2, the gains of Att.-HOD are better or close to the gains of Att. approach (except $|\mathcal{T}_{in}| = 2$).
                 
     \end{itemize}
\end{itemize}

In Figure \ref{fig:forecast}, we also provide sample forecasts comparing Att., Att.-HOD, Att.+PE and FC with the ground truth (GT) on an OOD profile indicating better generalization ability of Att.-HOD and Att.+PE, especially around points where Type-II load gets shifted. On the other hand, all methods perform well in IID setting as shown in Figure \ref{fig:iidforecast}.

\section{Conclusion and Future Work}
In this work, we consider the problem of demand response management from an electricity broker or retailer's perspective. We highlight temporal bias as an issue in optimizing profits via suitable tariff profile allocations. We motivate the need for better generalization to out-of-distribution profiles, and note that this is possible by leveraging the fact that consumers respond with same logic across profiles. We propose suitable inductive biases in deep neural networks-based approach for forecasting electricity consumption in response to new tariff profiles. This take the form of a permutation equivariance-enabled attention mechanism that can leverage the property of consumer behavior to respond in a certain way across profiles.
In future, it will be interesting to look at the generalization from the perspective of handling confounding bias as the historical profile allocation and the outcome are effected by the historical allocation policies, which in turn rely on the latent consumer attributes acting as confounders.

The current optimization objective takes into account broker's profit but ignores the cost of electricity for the end consumer - bringing this into the optimization objective is a potential next step.

\bibliography{aaai22}

\end{document}